%% file: root.tex
\documentclass[letterpaper, 10 pt, conference]{ieeeconf}  

\IEEEoverridecommandlockouts                              

\usepackage{soul}
\usepackage{xcolor}
\usepackage{cite}
\usepackage{import}
\usepackage{amsmath}
\usepackage{graphicx}
\usepackage{amsfonts} 

\title{\LARGE \bf
Trajectory Optimization for Quadruped Mobile Manipulators that Carry Heavy Payload
}

\author{Ioannis Dadiotis, Arturo Laurenzi and Nikos Tsagarakis
\thanks{This work was supported
by the European Union’s Horizon 2020 programme under Grant Agreements
No. 101016007 (CONCERT) and No. 871237 (SOPHIA).}
\thanks{All authors are with Humanoid and Human-Centered Mechatronics Research Line, Italian Institute of Technology, Genoa 16163, Italy {(emails: \tt\small name.surname@iit.it)}. Ioannis Dadiotis is also with
the Department of Informatics, Bioengineering, Robotics and Systems Engineering, University of Genoa, Genoa 16145, Italy.}
}

\usepackage{tikz}
\usepackage{textcomp}
\usepackage{lipsum}

\newcommand\copyrighttext{%
  \footnotesize \textcopyright 2022 IEEE. Personal use of this material is permitted.
  Permission from IEEE must be obtained for all other uses, in any current or future
  media, including reprinting/republishing this material for advertising or promotional
  purposes, creating new collective works, for resale or redistribution to servers or
  lists, or reuse of any copyrighted component of this work in other works.}
\newcommand\copyrightnotice{%
\begin{tikzpicture}[remember picture,overlay]
\node[anchor=south,yshift=10pt] at (current page.south) {\parbox{\dimexpr\textwidth-\fboxsep-\fboxrule\relax}{\copyrighttext}};
\end{tikzpicture}%
}

\begin{document}

\maketitle
\copyrightnotice

\thispagestyle{empty}
\pagestyle{empty}

\begin{abstract}

This paper presents a simplified model-based trajectory optimization (TO) formulation for motion planning on quadruped mobile manipulators that carry heavy payload of known mass. The proposed \emph{payload-aware} formulation simultaneously plans locomotion, payload manipulation and considers both robot and payload model dynamics while remaining computationally efficient. At the presence of heavy payload, the approach exhibits reduced leg outstretching (thus increased manipulability) in kinematically demanding motions due to the contribution of payload manipulation in the optimization. The framework's computational efficiency and performance is validated through a number of simulation and experimental studies with the bi-manual quadruped CENTAURO robot carrying on its arms a payload that exceeds 15 \% of its mass and traversing non-flat terrain.

\end{abstract}

\input{introduction}

\input{Trajectory_Optimization}

\input{WBC}

\input{Results}

\input{experiments}

\input{Conclusion}




\input{references}

\end{document}

%% file: introduction.tex
\section{INTRODUCTION}

Quadruped robots have in general outperformed wheeled platforms in rough terrain by taking advantage of their ability to discretely make and break contact with the environment. In contrast to aerial robots (which move without contact), they can, significantly, compensate and exert higher interaction forces due to the legged contact and articulation. This renders quadruped robots, compared to other mobile manipulation platforms, more promising for all-terrain applications that require executing manipulation actions with large physical interaction or increased payload capacity. This promise has in no case been fulfilled so far since very few works \cite{bellicoso2019alma, coros2021, bd_spot_arm, murphy2012high, yuntao2022, sleiman2021unified, chiu2022collisionfree, ferrolho2022roloma} have addressed the problem of simultaneously performing locomotion and manipulation tasks on real quadrupeds. 

\begin{figure}
  \centering
  \def\svgwidth{\columnwidth}
  \graphicspath{{figures/}}
  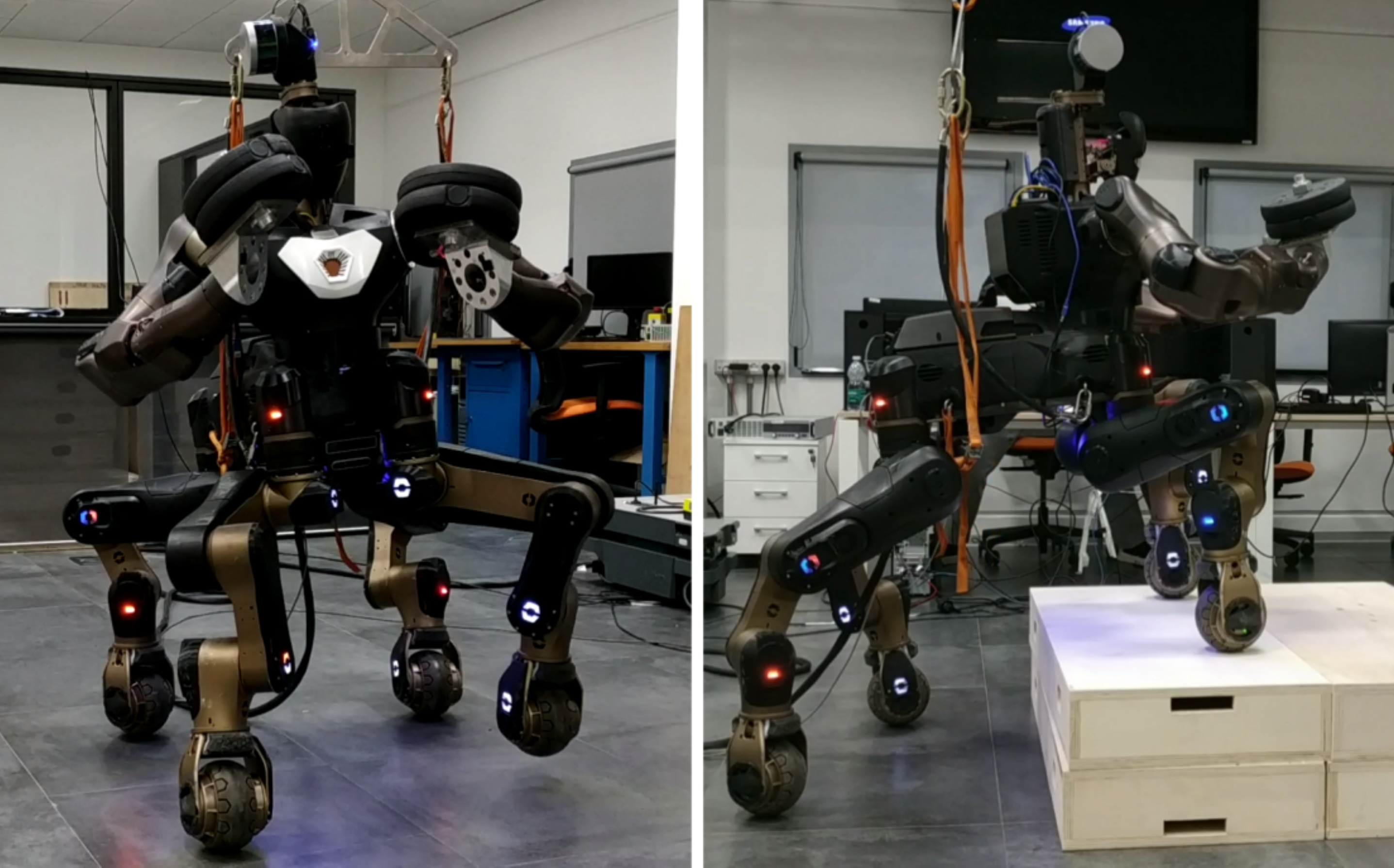
  \caption{The CENTAURO robot walking while carrying 17 kg payload. Corresponding video: https://youtu.be/09APxzIehpI}
  \label{Fig:robot_photo}
\end{figure}

Among other challenging tasks, quadruped manipulators promise robotizing heavy payload transportation in non-flat terrain, a task that is tedious for humans and unsuitable for their wheeled and aerial counterparts (while it has been also subject of research for humanoid robots \cite{humanoid_heavy_object}). At the presence of payload with substantial mass the robot dynamics are highly affected. 
As a result, locomoting while simultaneously carrying heavy payload poses significant challenges to the robot, namely compromising stability, forcing actuator saturation and reaching joint limits. 

Despite the existence of numerous quadruped manipulators \cite{bellicoso2019alma, coros2021, bd_spot_arm, rehman2018centaur, kashiri2019centauro, iit_teleoperativo, bd_dynamic_manipulation}, their deployment and feasibility to maintain locomotion while carrying heavy payload (more than 15 \% of the robot's mass) has been either relatively unexplored or compromised by generating motions only for the lower-body of the robot and overlooking the manipulation capabilities of the platform \cite{xinyuan2021}. Relying solely on locomotion for a task involving high interaction forces/payload at the robot's upper body can result in inefficient or at best suboptimal robot behavior. On the contrary, considering both locomotion and payload manipulation together at the planning stage is a more sophisticated alternative that can result in lower-body trajectories of greater margins, which can be tracked more easily. 

This work optimizes both locomotion and manipulation and considers both robot and payload simplified model dynamics together at the level of motion planning (before control). Deploying payload manipulation results in avoiding compromising kinematic performance of the robot’s lower-body and remaining far from leg singular configurations in kinematically demanding motions. The formulation is combined with a whole-body controller (WBC) and the framework, efficiently, generates a variety of motions (for flat and non-flat terrains) that are successfully executed in both simulation and real hardware under heavy payload.

\subsection{Related Work}
\label{Related Work}

Previous work on CENTAURO \cite{xinyuan2021} relies on total CoM estimation to plan only lower-body motion (no manipulation planning) for heavy payload transportation. Reaching the leg workspace kinematic limits (\emph{outstretching}) is traded off with zero moment point (ZMP) stability margin through a cost term. In contrast, the current approach leverages the platform's dual arm manipulation capabilities to avoid leg outstretching without compromising stability. Indeed, the framework achieves to plan and execute more challenging motions on both simulation and real hardware.

Existing works have explored separating locomotion and manipulation planning. In \cite{bellicoso2019alma} contact wrenches are commanded to the controller while the planner accounts only for locomotion. Similarly the BigDog robot of Boston Dynamics \cite{bd_dynamic_manipulation} was shown to carry and throw a cinder block \cite{murphy2012high}. The work of \cite{yuntao2022} proposes a learned locomotion policy that accounts for the predicted wrenches on the robot base planned by a separate model-based manipulation planner. The above approaches suffer from the same drawback in that locomotion planning is assigned to compensate for/reject the manipulation motion effect without having control over it. Thus, the manipulation behavior cannot be changed based on locomotion margins.

On the other hand locomotion and manipulation have been simultaneously optimized for bipeds, to name but a few \cite{dai2014whole, kudruss2015optimal}, and more recently for quadruped manipulators as follows. 
The full dynamics TO formulations of \cite{ferrolho, ferrolho2022roloma} have achieved robustness against payloads \cite{ferrolho2021residual} but are computationally expensive (prohibiting online planning for a robot much simpler than the one used in this paper). 

 The work of \cite{ewen2021} substitutes the feet contact forces with the ZMP for achieving tasks on flat terrain in simulation. On the contrary, the unified framework of \cite{sleiman2021unified} accounts for the centroidal robot dynamics and manipulated object to efficiently plan combined locomotion and manipulation tasks. Although the latter seems to find application for the heavy payload transportation task, scaling efficiently\footnote{In this paper, a TO is considered computationally efficient if the planning time is at least an order of magnitude shorter than the planning horizon.} to more complex robots as the CENTAURO (which consists of 39 actuated degrees of freedom [DoF] in contrast to the 16-DoF robot used) remains challenging\footnote{This is further compounded since by the time of writing the augmented-Lagrangian-based inequality constraint handling approach \cite{sleiman2021constraint} that is used in \cite{sleiman2021unified} is not yet open source released, thus cannot be off-the-shelf accessed.}. The current paper handles this complexity by adopting a middle ground between \cite{ewen2021} and \cite{sleiman2021unified} in terms of robot model descriptiveness. It accounts for a model richer than \cite{ewen2021} since optimizing for all contact forces as well as arm end-effectors (EE) motion. The resulted TO is computationally efficient (planning at least at 5 Hz with 4 sec. horizon) and is validated on the real CENTAURO robot, which is of much higher system dimension than any of the works above. Differently from other works, results and insight are presented from real experiments on non-flat terrain with 17 kg payload (85 \% of the arms' capacity and 15.1 \% of the robot mass). 

\subsection{Contribution}
\label{sec:Contribution}
Based on the current literature, as discussed in \ref{Related Work}, this work is characterized by the following contributions:
\begin{itemize}
\item A TO formulation for motion planning of quadruped manipulators carrying heavy payload. The formulation simultaneously plans locomotion and payload manipulation and as a result, in contrast to locomotion-only, avoids excessive lower-body motion that can cause leg outstretching (boundary singularities) when stepping.
\item The formulation is used for (but, as shown in Sec. \ref{sec:computational}, not restricted to) offline planning. Combined with a WBC it is evaluated in various simulation and experimental scenarios with the CENTAURO robot \cite{kashiri2019centauro}, demonstrating the ability to generate motions that can be tracked by real robots of high complexity. The obstacles negotiated in both simulation and real experiment have not been shown before from a quadruped manipulator that carries such heavy payload.
\item The approach is computationally efficient for high dimensional robots (as is a bi-manual quadruped) such that can be straightforwardly extended to online receding horizon TO. This work is the first to provide insights on the application and efficiency of simplified model-based TO on quadrupeds with more than one arms. 
\end{itemize}

%% file: figures/different_robot_photo.pdf_tex
\begingroup%
  \makeatletter%
  \providecommand\color[2][]{%
    \errmessage{(Inkscape) Color is used for the text in Inkscape, but the package 'color.sty' is not loaded}%
    \renewcommand\color[2][]{}%
  }%
  \providecommand\transparent[1]{%
    \errmessage{(Inkscape) Transparency is used (non-zero) for the text in Inkscape, but the package 'transparent.sty' is not loaded}%
    \renewcommand\transparent[1]{}%
  }%
  \providecommand\rotatebox[2]{#2}%
  \newcommand*\fsize{\dimexpr\f@size pt\relax}%
  \newcommand*\lineheight[1]{\fontsize{\fsize}{#1\fsize}\selectfont}%
  \ifx\svgwidth\undefined%
    \setlength{\unitlength}{828.0000173bp}%
    \ifx\svgscale\undefined%
      \relax%
    \else%
      \setlength{\unitlength}{\unitlength * \real{\svgscale}}%
    \fi%
  \else%
    \setlength{\unitlength}{\svgwidth}%
  \fi%
  \global\let\svgwidth\undefined%
  \global\let\svgscale\undefined%
  \makeatother%
  \begin{picture}(1,0.62228261)%
    \lineheight{1}%
    \setlength\tabcolsep{0pt}%
    \put(0,0){\includegraphics[width=\unitlength,page=1]{different_robot_photo.pdf}}%
  \end{picture}%
\endgroup%

%% file: Trajectory_Optimization.tex
\section{TRAJECTORY OPTIMIZATION FORMULATION}
\label{Sec: TO_main}

In this work, the motion planning problem is formulated using direct transcription/collocation which transcribes the continuous optimization problem in a constrained Nonlinear Programming (NLP) problem. This is done by discretizing the horizon of the optimization in a number of knots and the solver is assigned to find the optimal values of the variables at these knots. The optimal values of the obtained solution are then interpolated. Although many off-the-shelf NLP solvers exist interior-point ones are faster \cite{ferrolho2021residual}, \cite{pardo2016evaluating} and can handle any type of constraint, thus, they are preferred.

The TO formulation plans both locomotion and manipulation trajectories for the task of carrying heavy payload with known mass\footnote{The mass of a grasped payload can be estimated through force estimation at the EE using a wrist force/torque sensor or by exploring the joint torque sensing available on the arm.}, thus referred as \emph{payload-aware}. It consists of a CoM and arm EE motion planning framework, i.e. feet trajectories are not optimized. The latter are, heuristically, planned before the TO based on user inputs (gait pattern, stride length and duration, total time $T$ of the motion and step vertical clearance). Based on the above, feet EE trajectories are known and introduced in the TO as NLP parameters.

The robot arms are considered in rigid (prehensile) contact with the payload and able to manipulate it, thus, the motion of the payload is identical to the motion of the arm EE. The CENTAURO robot has two arms and, henceforth, this paper focuses on scenarios with one payload at each arm. However the approach can be easily adapted for different quadrupeds. For the rest of this paper all position vectors are expressed in a fixed inertial (world) frame whose notation is omitted.

\subsection{Decision variables}
\label{sec:decision_Variables}

The formulation optimizes the CoM state $\pmb{z}(t) = \begin{bmatrix}\pmb{r}(t)& \dot{\pmb{r}}(t)& \ddot{\pmb{r}}(t)\end{bmatrix}^T$, which includes position, velocity and acceleration vectors, the CoM jerk $\dddot{\pmb{r}}(t)$, the motion of the arm EEs as well as the forces $\pmb{f}_i(t)$ at all (feet and arm) EEs. Fixed-step discretization is used, thus $t=k \cdot \Delta t$ where $k \in \{0,...,N-1\}$ and $N$ the number of knots. For the rest of this paper the time dependency of the decision variables is omitted. The CoM position and EE forces are interpolated as cubic splines and piecewise linear, respectively. Arm EE position trajectories are parameterized as cubic splines based on Cubic Hermite Parameterization (CHP) \cite{winklergait}.

The initial and final value of some of the decision variables is specified with equality constraints on the first and last NLP knot (initial and final conditions, respectively). In particular, before the beginning of each TO the CoM and arm EEs position are perceived and enforced. All decision variables related with velocity and acceleration are enforced to be zero at both the beginning and the end of the TO so that the robot starts and stops smoothly. The final CoM position is bounded within a desired region (centered around the nominal CoM position consistent with the final footholds) through an inequality constraint. 

\subsection{The robot model}
\begin{figure}
  \centering
  \def\svgwidth{\columnwidth}
  \graphicspath{{figures/}}
  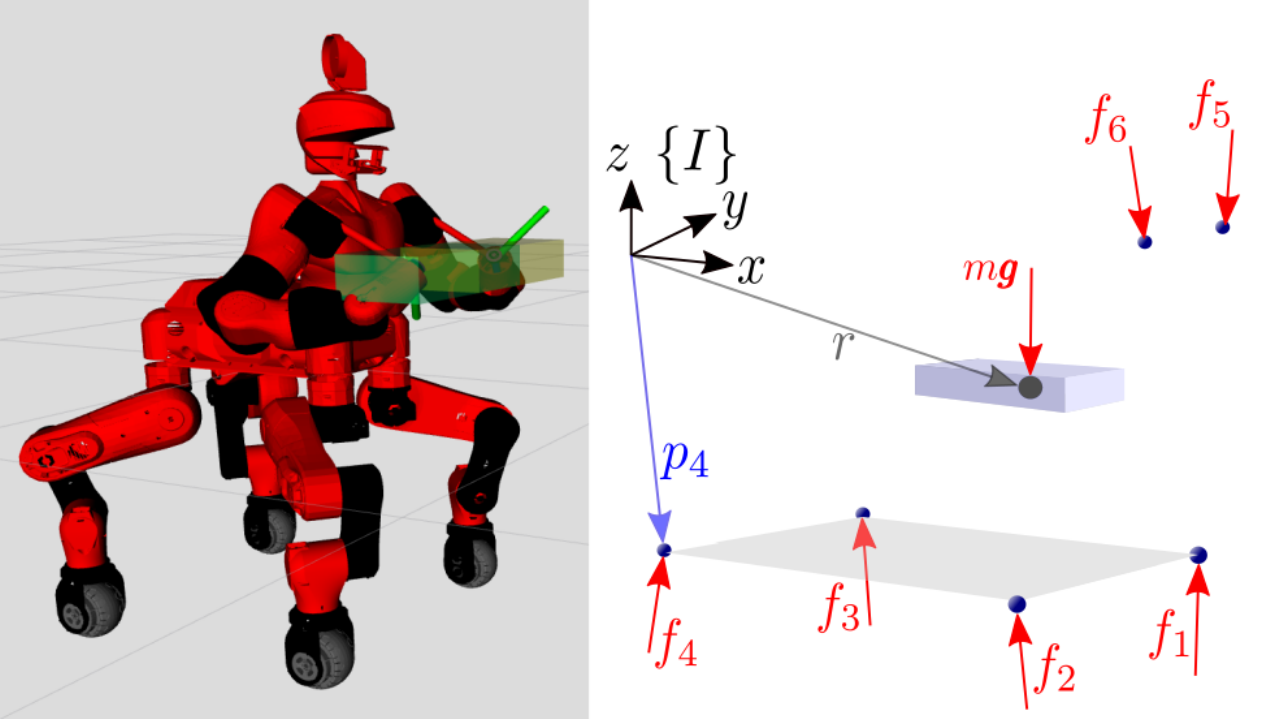
  \caption{Visualization of the CENTAURO robot with the arm EE box workspaces (left). The simplified dynamic model considered by the payload-aware planner (right). For the sake of clarity, the position vector $\pmb{p}_i$ with respect to (wrt) the inertial frame $I$ is depicted for one EE.}
  \label{Fig:Model}
\end{figure}

The robot is modeled using the Single Rigid Body Dynamics (SRBD) model, which assumes rigid robot links, negligible momentum produced by the joint velocities and that full-body inertia remains similar to the one in nominal joint position. Additionally, point contacts are assumed. The SRBD model is described by the following equation:
\begin{align}
    \begin{bmatrix}
        \dot{\pmb{P}}\\
        \dot{\pmb{L}}
    \end{bmatrix}
    = \begin{bmatrix}
        m \pmb{g} + \sum_{i=1}^{n_{cont}}{\pmb{f}_i}\\
        \sum_{i=1}^{n_{cont}}{(\pmb{p}_i - \pmb{r}) \times \pmb{f}_i}
    \end{bmatrix}
    \label{eq:dynamics}
\end{align}
where $\dot{\pmb{P}} = m \cdot \ddot{\pmb{r}}$ and $\dot{\pmb{L}}$ are the derivatives of the linear and angular momentum, respectively, $\pmb{g}$ is the gravity vector, $\pmb{p}_i$ denotes the feet and arm EE position vectors, $m$ is the robot mass and the number of contacts $n_{cont}=6$ (considering that both feet and arm EEs make contact with the environment/payload). Base angular motion is not optimized (it is generated by the WBC of Sec. \ref{sec:WBC}) and constant angular momentum is assumed $\dot{\pmb{L}} = \pmb{0}$ so as to maximize efficiency. An illustration of the model is shown in Fig. \ref{Fig:Model}.

\subsection{Locomotion-related behavior}
\label{sec:locomotion}



During legged locomotion the robot has to maintain stability/balance and not slip on the ground. These conditions are satisfied with constraints related with the feet EE contact forces, namely the stability constraint \eqref{eq:unilateral} and friction pyramid \eqref{eq:friction} (preferred over the second-order cone due to linearity) constraints.
\begin{align}
    \pmb{f}_i^T \pmb{\hat{n}}_i &\geq f^z_{min} \label{eq:unilateral}\\
    |\pmb{f}_i^T \pmb{\hat{t}}_i^j| &\leq  \mu \cdot \pmb{f}_i^T \pmb{\hat{n}}_i \label{eq:friction}
\end{align}
where $j=\{x,y\}$, $\pmb{\hat{n}}_i$, $\pmb{\hat{t}}_i^x$, $\pmb{\hat{t}}_i^y$ denote the unit vectors normal and parallel to the tangential contact plane of foot $i$ and $\mu$ is the friction coefficient. Large stability bounds are imposed by setting a positive lower bound of $f^z_{min}=100$ N (9 \% of the robot's weight) in \eqref{eq:unilateral} for the normal component of the force at each foot in contact. The above positive bound was selected from trials in simulation and is necessary in order to compensate for the SRBD model assumptions, like the neglect of the momentum produced by the joint velocities which is significant due to the considerable distal mass of CENTAURO (leg to robot mass ratio is 10.8 \%). This way the stance feet do not lose contact with the terrain and the robot remains stable. For each swinging foot, the corresponding force is set to zero and, thus, is not optimized.

CoM jerk and feet force components $\pmb{f}_i^{xy}$ (where $i \in \{0,...,4\}$) are penalized to avoid oscillatory trajectories and favor forces close to contact normals, respectively. Finally, a penalty cost is included for the CoM position with the form:
\begin{equation}
    J_r = \| \pmb{r} - (\pmb{p_{mean}} + \pmb{c_{ref}})\|^2
    \label{eq:com_ref_cost}
\end{equation}
 where $n_{feet}=4$, $\pmb{p_{mean}}=\frac{1}{n_{feet}}\sum_{i=1}^{n_{feet}} \pmb{p}_i$ is the mean of the feet EE position vectors (regardless of contact state) and $\pmb{c_{ref}}$ is a robot-specific vector with only vertical component (so that the CoM reference point is above $\pmb{p_{mean}}$). This way the horizontal deviation of the CoM from the geometric center of the polygon formed by the feet EEs is penalized.
 
It is worth mentioning that the above locomotion-related elements of the formulation can make up alone a reliable formulation for planning locomotion (i.e. without moving the arms wrt the robot base). This can be done by excluding the two arm EEs from the robot model \eqref{eq:dynamics}, the manipulation-related decision variables (arm EE trajectories and forces of Sec. \ref{sec:decision_Variables}) as well as any manipulation cost/constraint (elements of Sec. \ref{sec:manipulation}). Such a locomotion-only formulation is used for comparison in Sec. \ref{sec:results_main}, the competitiveness of which is thoroughly explained in that section. 

\subsection{Payload manipulation-related behavior}
\label{sec:manipulation}

Similar to \cite{sleiman2021unified}, the motion planner accounts for the payload dynamics. The motivation for this is that the payload dynamics are directly interacting with the robot dynamics through the interaction forces at the arm EEs. Here, each payload is modeled as a point mass to which a force $\pmb{f}_{\pmb{pay}, i}$ is exerted by the grasping robot arm. Therefore, its dynamics can be described with the Newton's equation of motion \eqref{eq:payload_dynamics}:
\begin{equation}
    m_{pay} \cdot \pmb{\ddot{p}_{pay}} = m_{pay} \cdot \pmb{g} + \pmb{f}_{\pmb{pay}, i} ,\hspace{0.5cm} i \in \{5,6\}
    \label{eq:payload_dynamics}
\end{equation}
where $\pmb{\ddot{p}_{pay}} = \ddot{\pmb{p}}_{i}$ is the payload acceleration (identical to the i-th EE), $m_{pay}$ is the payload mass and $\pmb{f}_{\pmb{pay}, i} = -\pmb{f}_i$ is equal and opposite to the arm EE force.

The arm EE trajectories should remain within their kinematic range without self-colliding and, thus, the solver is constrained to do so. Box constraints are preferred because of convexity and linearity. The workspace of each arm EE (shown in Fig. \ref{Fig:Model}) is centered at the nominal position wrt the CoM $\bar{\pmb{p}}_{ri}$ and aligned with the inertial frame.  Since base orientation is not available at the planner (it is not optimized) boxes cannot be expressed wrt the base frame. For the variety of scenarios presented in this work this is not a limitation, however it may be crucial for more complex maneuvers. The constraint for each arm EE can be described as:
\begin{align}
    \label{eq:box_ee}
    |(\pmb{p}_i - \pmb{r} - \bar{\pmb{p}}_{ri})^T   \pmb{\hat{j}}| \leq 0.5 \cdot \pmb{b_{ee}}^T   \pmb{\hat{j}}
\end{align}
where $\pmb{\hat{j}} \in \{\pmb{\hat{x}}, \pmb{\hat{y}}, \pmb{\hat{z}}\}$ are the unit vectors along the inertial directions and $\pmb{b_{ee}}$ is a $1 \times 3$ array matrix including the box dimensions. The two workspaces are selected to overlap with each other, as shown in Fig. \ref{Fig:Model}, in order to provide the solver with more freedom regarding the arm motion. The bigger the workspaces are, the more freedom the solver has to manipulate the payload. Due to this overlap a constraint is added to ensure no self-collision by keeping the distance between the two EEs along the $y$ inertial direction greater than a safety threshold $b_s > 0$, as shown in \eqref{eq:self_collision}. 
\begin{align}
    \label{eq:self_collision}
    |(\pmb{p}_5 - \pmb{p}_6)^T \pmb{\hat{y}}| \ \geq b_s
\end{align}

Finally, the magnitude of each arm EE force is bounded with a box constraint, which results in bounding the corresponding arm EE acceleration. This is because each arm EE and the corresponding payload are subject to the same acceleration, which is coupled with the corresponding arm EE force through the payload dynamics (\ref{eq:payload_dynamics}). Each arm EE force box is centered at the payload's weight vector $m_{pay}\cdot\pmb{g}$, thus the constraint has the form:
\begin{align}
    | (\pmb{f}_i - m_{pay} \cdot \pmb{g})^T \pmb{\hat{j}} | & \leq 0.5 \cdot \pmb{b_f}^T \pmb{\hat{j}}
    \label{eq:bound_arm_force}
\end{align}
where $\pmb{b_f}$ is a $1 \times 3$ array matrix including the box dimensions. The larger the bounding boxes are, the more dynamic arm motions the framework is permitted to plan. For the CENTAURO robot $\pmb{b_f} = \begin{bmatrix}16 & 16 & 6\end{bmatrix} N$ was found to work well. Based on \eqref{eq:payload_dynamics} this results in admissible accelerations up to $\begin{bmatrix}0.8 & 0.8 & 0.3\end{bmatrix} m/s^2$ (for the three inertial directions, respectively) for a 10 kg payload. 

\begin{figure}[ht]
  \centering
  \def\svgwidth{\columnwidth}
  \graphicspath{{figures/}}
  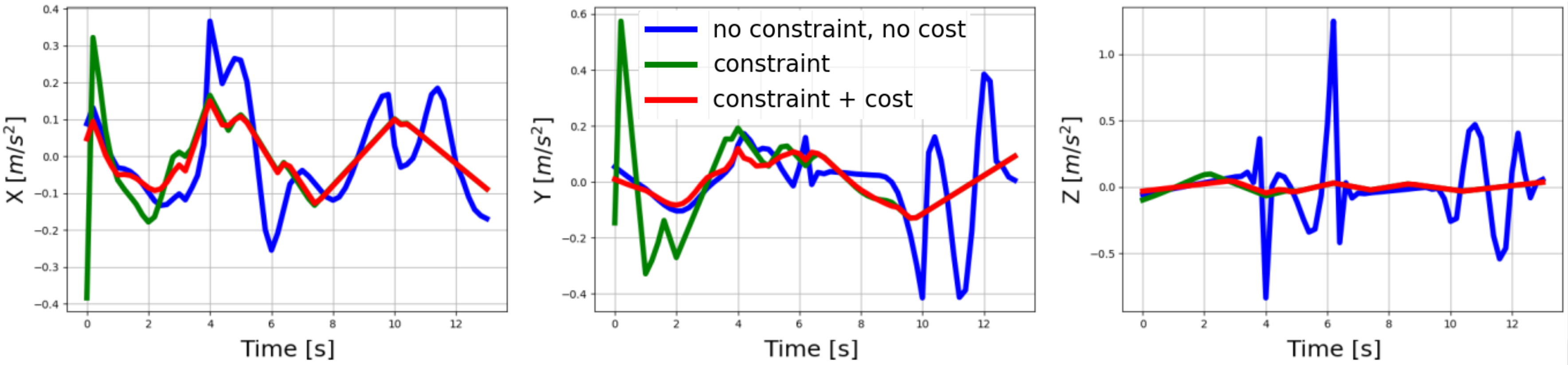
  \caption{Left arm EE acceleration (plans from \emph{Sc. 2} of Section \ref{subsec:singularities}). Constraint \eqref{eq:bound_arm_force} and cost \eqref{eq:analytical_cost} result in motions with reduced arm EEs acceleration.}
  \label{Fig:analytical_cost}
\end{figure}

Penalty terms related to the desired manipulation behavior are also considered. A large penalty cost is added at the last NLP knot in order to favor final arm EEs position close to nominal. For each arm a cost is added to favor motions with small EE acceleration (among the ones specified through \eqref{eq:bound_arm_force}). This cost has analytical form and penalizes the integral of the squared acceleration polynomial, as shown in \eqref{eq:analytical_cost}.
\begin{equation}
    J_i = \sum_{k=0}^{N-2}{\int_{k\cdot \Delta t}^{(k+1)\cdot \Delta t} \pmb{\ddot{p}}_{\pmb{poly}, i}^2(t) \,dt}
    \label{eq:analytical_cost}
\end{equation}
where $\pmb{\ddot{p}}_{\pmb{poly}, i}(t)$ is the arm EE optimized acceleration polynomial (which is a function of the decision variables at the adjacent knots according to CHP \cite{winklergait}), $\Delta t$ is the duration of each time segment and $N$ the number of the time segments according to the NLP discretization. The analytical cost penalizes acceleration through the whole spline and not just at the knots. An example of the effect of cost \eqref{eq:analytical_cost} as well as constraint \eqref{eq:bound_arm_force} on the acceleration of the arm EEs is shown in Fig. \ref{Fig:analytical_cost}.

The presented formulation can be easily adapted to quadrupeds with different number of arms by enforcing all constrains of this chapter, except from \eqref{eq:self_collision}, for each arm EE. Constraint \eqref{eq:self_collision} should be imposed for every pair of arms while vanishes for a single arm.


%% file: figures/model_comp.pdf_tex
\begingroup%
  \makeatletter%
  \providecommand\color[2][]{%
    \errmessage{(Inkscape) Color is used for the text in Inkscape, but the package 'color.sty' is not loaded}%
    \renewcommand\color[2][]{}%
  }%
  \providecommand\transparent[1]{%
    \errmessage{(Inkscape) Transparency is used (non-zero) for the text in Inkscape, but the package 'transparent.sty' is not loaded}%
    \renewcommand\transparent[1]{}%
  }%
  \providecommand\rotatebox[2]{#2}%
  \newcommand*\fsize{\dimexpr\f@size pt\relax}%
  \newcommand*\lineheight[1]{\fontsize{\fsize}{#1\fsize}\selectfont}%
  \ifx\svgwidth\undefined%
    \setlength{\unitlength}{365.66929134bp}%
    \ifx\svgscale\undefined%
      \relax%
    \else%
      \setlength{\unitlength}{\unitlength * \real{\svgscale}}%
    \fi%
  \else%
    \setlength{\unitlength}{\svgwidth}%
  \fi%
  \global\let\svgwidth\undefined%
  \global\let\svgscale\undefined%
  \makeatother%
  \begin{picture}(1,0.56589147)%
    \lineheight{1}%
    \setlength\tabcolsep{0pt}%
    \put(0,0){\includegraphics[width=\unitlength,page=1]{model_comp.pdf}}%
  \end{picture}%
\endgroup%

%% file: figures/sc2_leftarm_acc_cost.pdf_tex
\begingroup%
  \makeatletter%
  \providecommand\color[2][]{%
    \errmessage{(Inkscape) Color is used for the text in Inkscape, but the package 'color.sty' is not loaded}%
    \renewcommand\color[2][]{}%
  }%
  \providecommand\transparent[1]{%
    \errmessage{(Inkscape) Transparency is used (non-zero) for the text in Inkscape, but the package 'transparent.sty' is not loaded}%
    \renewcommand\transparent[1]{}%
  }%
  \providecommand\rotatebox[2]{#2}%
  \newcommand*\fsize{\dimexpr\f@size pt\relax}%
  \newcommand*\lineheight[1]{\fontsize{\fsize}{#1\fsize}\selectfont}%
  \ifx\svgwidth\undefined%
    \setlength{\unitlength}{1018.16548505bp}%
    \ifx\svgscale\undefined%
      \relax%
    \else%
      \setlength{\unitlength}{\unitlength * \real{\svgscale}}%
    \fi%
  \else%
    \setlength{\unitlength}{\svgwidth}%
  \fi%
  \global\let\svgwidth\undefined%
  \global\let\svgscale\undefined%
  \makeatother%
  \begin{picture}(1,0.23280457)%
    \lineheight{1}%
    \setlength\tabcolsep{0pt}%
    \put(0,0){\includegraphics[width=\unitlength,page=1]{sc2_leftarm_acc_cost.pdf}}%
  \end{picture}%
\endgroup%

%% file: WBC.tex
\section{WHOLE-BODY CONTROLLER (WBC)}
\label{sec:WBC}

\begin{figure*}
  \centering
  \def\svgwidth{\textwidth}
  \graphicspath{{figures/}}
  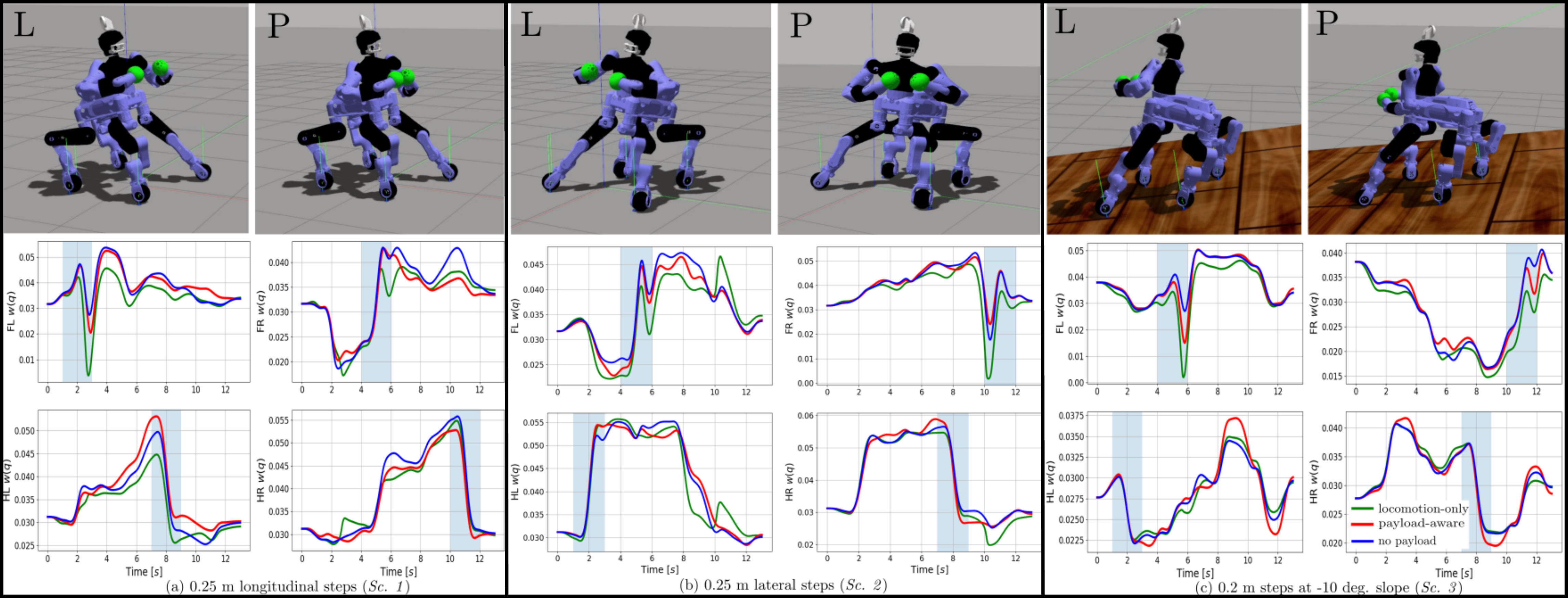
  \caption{The robot during the execution of simulation experiments with 10 kg payload at each arm (top). Snaphots are showed for both locomotion-only (L) and payload-aware (P) cases for Sc. 1-3. The attached green ball EEs consist the payloads. Manipulability metric of each leg for the locomotion-only, payload-aware and no payload cases for Sc. 1-3 (bottom). The shaded regions denote swing periods.}
  \label{Fig:simulations}
\end{figure*}

Motion generation requires specifying whole-body trajectories and, thus, a WBC based on hierarchical optimization and inverse kinematics (using RBDL library \cite{Felis2016}) is developed within the framework of \cite{hoffman2017robot}. The WBC accepts the CoM and EE position plans and generates whole-body joint position trajectories which are fed to the low-level joint position controllers through dedicated ROS-based software \cite{laurenzi2019cartesi, muratore2017xbotcore}. The structure of the WBC (stack of tasks) is shown in Table \ref{tab:wbc_structure}. It is worth mentioning that sufficient control authority is assumed and joint torque limits are not considered which, nevertheless, does not diminish the execution of the variety of motions presented in this work.

\begin{table}[ht]
\caption{WBC tasks and constraints}
\label{tab:wbc_structure}
\begin{center}
\begin{tabular}{|c||c|}
\hline
Priority & Tasks\\
\hline
1 & Feet EE position tracking\\
\hline
1 & CoM position tracking\\
\hline
2 & Arms EE position tracking\\
\hline
3 & Postural task\\
\hline
Constraint & Joint limits\\
\hline
Constraint & Velocity limits\\
\hline
\end{tabular}
\end{center}
\end{table}

%% file: figures/NEW_simulations_comp.pdf_tex
\begingroup%
  \makeatletter%
  \providecommand\color[2][]{%
    \errmessage{(Inkscape) Color is used for the text in Inkscape, but the package 'color.sty' is not loaded}%
    \renewcommand\color[2][]{}%
  }%
  \providecommand\transparent[1]{%
    \errmessage{(Inkscape) Transparency is used (non-zero) for the text in Inkscape, but the package 'transparent.sty' is not loaded}%
    \renewcommand\transparent[1]{}%
  }%
  \providecommand\rotatebox[2]{#2}%
  \newcommand*\fsize{\dimexpr\f@size pt\relax}%
  \newcommand*\lineheight[1]{\fontsize{\fsize}{#1\fsize}\selectfont}%
  \ifx\svgwidth\undefined%
    \setlength{\unitlength}{1319.8346572bp}%
    \ifx\svgscale\undefined%
      \relax%
    \else%
      \setlength{\unitlength}{\unitlength * \real{\svgscale}}%
    \fi%
  \else%
    \setlength{\unitlength}{\svgwidth}%
  \fi%
  \global\let\svgwidth\undefined%
  \global\let\svgscale\undefined%
  \makeatother%
  \begin{picture}(1,0.38117247)%
    \lineheight{1}%
    \setlength\tabcolsep{0pt}%
    \put(0,0){\includegraphics[width=\unitlength,page=1]{NEW_simulations_comp.pdf}}%
  \end{picture}%
\endgroup%

%% file: Results.tex
\section{RESULTS AND EVALUATION}
\label{sec:results_main}

This section presents results from the motions generated by the payload-aware formulation of Sec. \ref{Sec: TO_main} combined with the WBC of Sec. \ref{sec:WBC}. The planner is compared with the case of considering payload as part of the robot model (as a fully observable link) and using locomotion-only planning. More specifically, this is derived by excluding all the manipulation-related elements (i.e. the two arm EEs in the robot model \eqref{eq:dynamics}, the decision variables of arm EE motion and forces as well as all costs/constraints of Sec. \ref{sec:manipulation}).
 This planner is, henceforth, referred as \emph{locomotion-only} case.
The motivation for this comparison is that the locomotion-only case was found to be more efficient than our previous framework \cite{xinyuan2021} (which is a state-of-the-art method handling such heavy payload), achieving larger strides and traversing larger gaps and obstacles. The underlying reasons are that \cite{xinyuan2021} accounts for a less descriptive (linear) dynamic model and restricts the orientation of the robot base (at the WBC stage) to be horizontal, thus angular base motion is not contributing to CoM motion which is rather conservative for robots with upper body/arms. Finally, \cite{xinyuan2021} is based on total CoM estimation that comes with estimation errors while the locomotion-only case in our comparisons considers perfect knowledge of the payload (which can be easily applied in simulation). Based on the above, the locomotion-only case consists a more competitive framework which handles heavy payload of known mass and, thus, is used for the following comparisons.  
For the sake of completeness the case of planning only locomotion (again by excluding the manipulation-related elements) without any payload carried by the robot, named as \emph{no payload} case, is also included in the comparisons. The same WBC is used in all cases. 

The comparisons presented in this section are based on three simulated scenarios\footnote{All motions can be found in the submitted video which is also available on https://youtu.be/09APxzIehpI .} \emph{(Sc.)} while the robot carries 10 kg payload at each arm (full payload capacity for each arm and total 17.8 \% of the robot's mass):\\
\emph{Sc. 1)} 4 longitudinal steps of 0.25 m on flat terrain. Due to the asymmetrical wrt the lateral axis (forward oriented) robot and grasped payload mass distribution such large strides are kinematically demanding for the front robot legs. Locomotion-only exhibits excessive backward base motion resulting in front leg configurations close to singularities.\\
\emph{Sc. 2)} 4 lateral steps of 0.25 m on flat terrain. This motion highlights that the proposed approach, also, overcomes locomotion-only planning in large lateral strides, despite the symmetrical robot and payload mass distribution wrt the longitudinal axis.\\
\emph{Sc. 3)} 4 steps of 0.2 m on a -10 degrees inclined terrain. This scenario is more challenging than Sc. 1 since the negative slope necessitates shifting the CoM more backwards\footnote{On flat terrain, the robot loses static stability when the CoM projection on the ground along the gravity vector exits the support polygon \cite{wieber2016modeling}.} and highlights the efficiency of the payload-aware approach in traversing sloped terrain under heavy payload. All motions of Sc. 1-3 have 13 sec. duration.

\subsection{Kinematically demanding motions}
\label{subsec:singularities}

Unless very dynamic, quadrupeds have to move their CoM inward the support polygon before step lift-off to maintain stability. Under heavy payload, larger CoM motions may be necessitated in order to compensate for the payload effect. In kinematically demanding motions, e.g. large strides where swing distance is large, swing leg may be outstretched and reach its workspace kinematic limit, a configuration known as \emph{boundary singularity}.
In this work, metric \eqref{eq:manipulability_metric} is used to evaluate the distance of a leg configuration from this kind of singularity.
\begin{equation}
    w(\pmb{q}) = \sqrt{\det(\pmb{J_u}(\pmb{q})\pmb{J_u}^T(\pmb{q}))}
    \label{eq:manipulability_metric}
\end{equation}
where $\pmb{J_u}$ is the linear velocity jacobian of the foot EE wrt the base link while the joint position vector $\pmb{q}$ is computed by the WBC. This metric is proportional to the linear velocity manipulability ellipsoid and the closer its value is to zero, the closer the configuration is to a singularity.

In Fig. \ref{Fig:simulations}, simulation snapshots and metric \eqref{eq:manipulability_metric} for each leg during Sc. 1-3 are depicted for the payload-aware and locomotion-only cases. In Sc. 1, 3 the locomotion-only motion plans (green color in Fig. \ref{Fig:simulations}) result in a front left (FL) leg configuration close to singularity before touchdown. This kinematic inefficiency is becoming evident at smaller steps in Sc. 3 than Sc. 1 because of the negative slope. Additionally, during Sc. 2 forward right (FR) leg reaches a configuration close to singularity before lift-off, since the robot base is moving inwards the future support polygon. On the contrary, in all above scenarios the payload-aware planner generates motions that are singularity free, as is the case for the no payload case. Even more, in Sc. 2 payload-aware planner manages to exceed the no payload case in terms of FR leg manipulability. Based on the above comparisons, although manipulability is not explicitly considered in the formulation (as a constraint or cost), avoiding excessive leg outstretching emerges naturally from engaging payload manipulation planning that contributes to constraint satisfaction (due to the substantial payload mass). As a result the framework provides larger freedom for shaping CoM trajectories (and, thus, lower-body motion). Finally, the above strides were indeed tried and found infeasible using our previous framework \cite{xinyuan2021}. 

\subsection{Payload manipulation contribution}
\label{subsec:results_manipulation}

\begin{figure}
  \centering
  \def\svgwidth{\columnwidth}
  \graphicspath{{figures/}}
  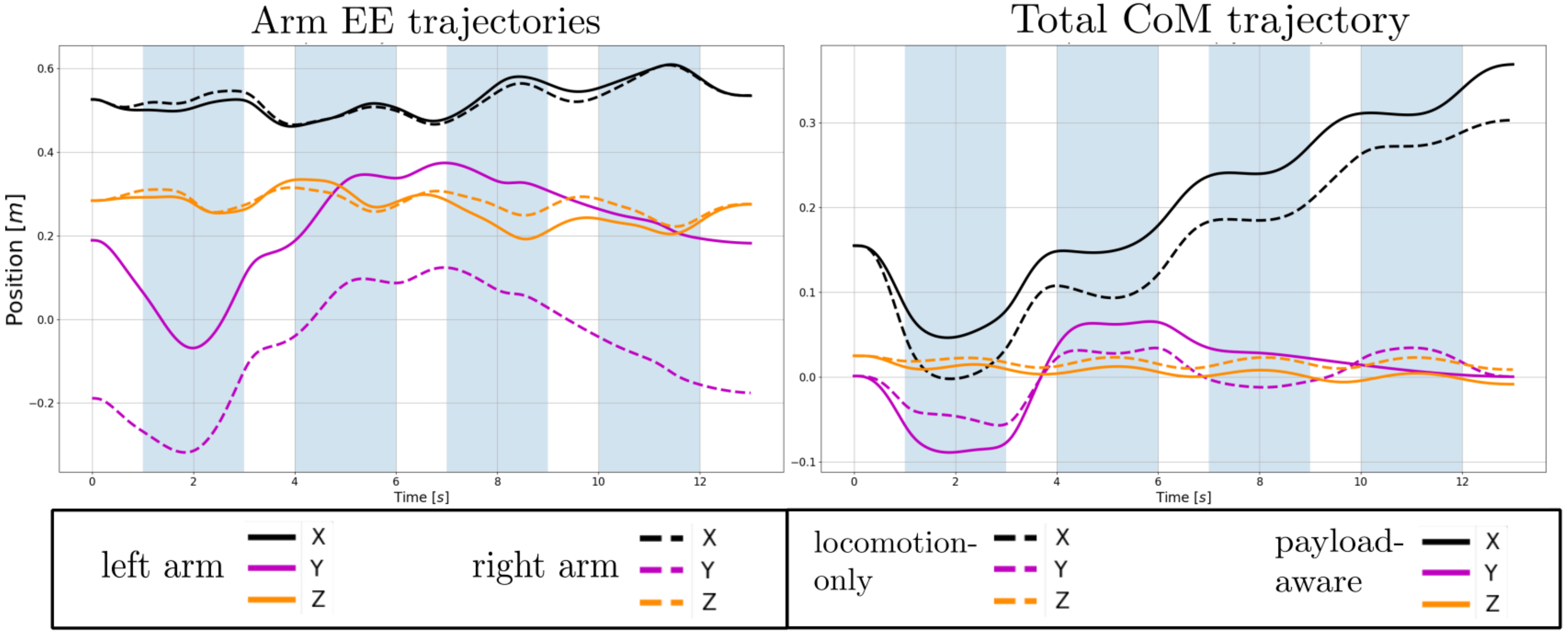
  \caption{Arm EE trajectories wrt to base link planned by the payload-aware planner (left). Comparison of total CoM position for the payload-aware and locomotion-only for Sc. 1 (right). The same trends were observed for Sc. 2-3 (omitted due to lack of space).}
  \label{Fig:arm_ee_trj}
\end{figure}

\begin{figure}
  \centering
  \def\svgwidth{\columnwidth}
  \graphicspath{{figures/}}
  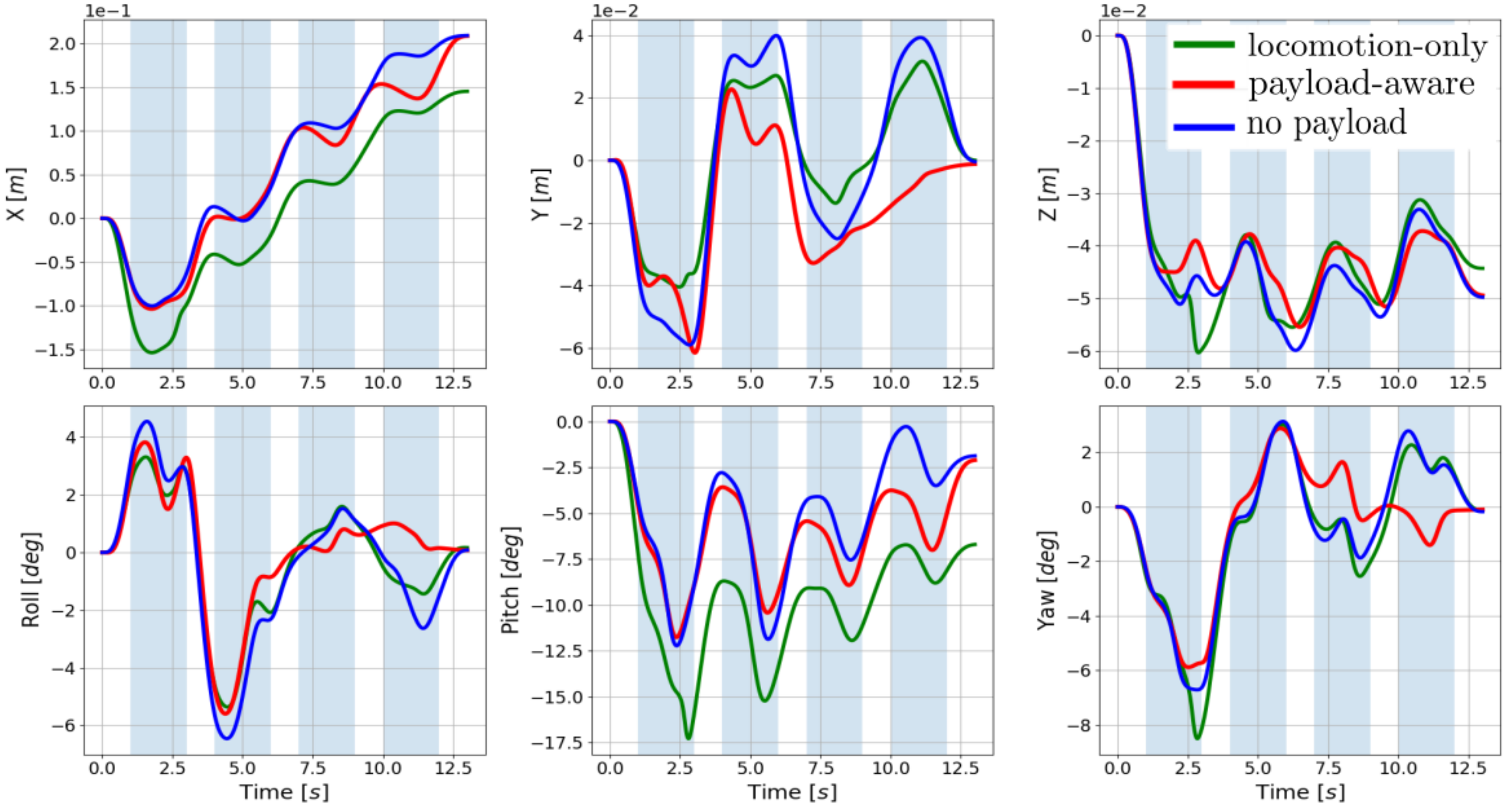
  \caption{Comparison of the 6 DoF base trajectories for the locomotion-only, payload-aware and no payload cases during Sc. 1. The same trends were observed for Sc. 2-3 (omitted due to lack of space).}
  \label{Fig:pelvis_trj}
\end{figure}

Fig. \ref{Fig:arm_ee_trj} (left) shows the arm EE trajectories for the payload-aware planner wrt the base link for Sc. 1. The graph demonstrates the importance of the arm EE motion, especially in the $y$ direction (with a span of more than 0.4 m in some cases) due to the selected boxes in constraint \eqref{eq:box_ee}. Moreover, the effect of constraint \eqref{eq:self_collision} is evident from the maintained distance between the two EEs in the $y$ direction.

Fig. \ref{Fig:arm_ee_trj} (right) depicts the total (robot and payload) CoM trajectory of locomotion-only, and payload-aware cases for Sc. 1. Overall, in the payload-aware case the total CoM demonstrates a clear tendency to move less backward and more on the lateral direction compared to the locomotion-only. This preference in lateral motions is due to the larger freedom provided in that direction by our formulation.

As shown in Fig. \ref{Fig:pelvis_trj} the tendency of the locomotion-only case to move the total CoM more backwards results in a similar base link motion, since the payload mass is not manipulated. Additionally, the base exhibits significantly larger pitch up motions which are, also, observed in top Fig. \ref{Fig:simulations} (the difference reaches 7 degrees). Pitch motion is generated by the WBC's effort to track the CoM longitudinal plans\footnote{Since the total CoM is forward concentrated pitch motion contributes in moving the CoM backwards.}. On the contrary, although the total CoM in the payload-aware case follows large lateral motions, this does not affect the motion of the base to a large extent due to the contribution of the payload manipulation in this direction (separately from the robot's CoM). Therefore, leveraging payload manipulation results in avoiding excessive linear and angular base motions and reduced lower-body motions.

\subsection{Negotiating non-flat terrain}
\label{sec:non_flat_terrain}

\begin{figure}
  \centering
  \def\svgwidth{\columnwidth}
  \graphicspath{{figures/}}
  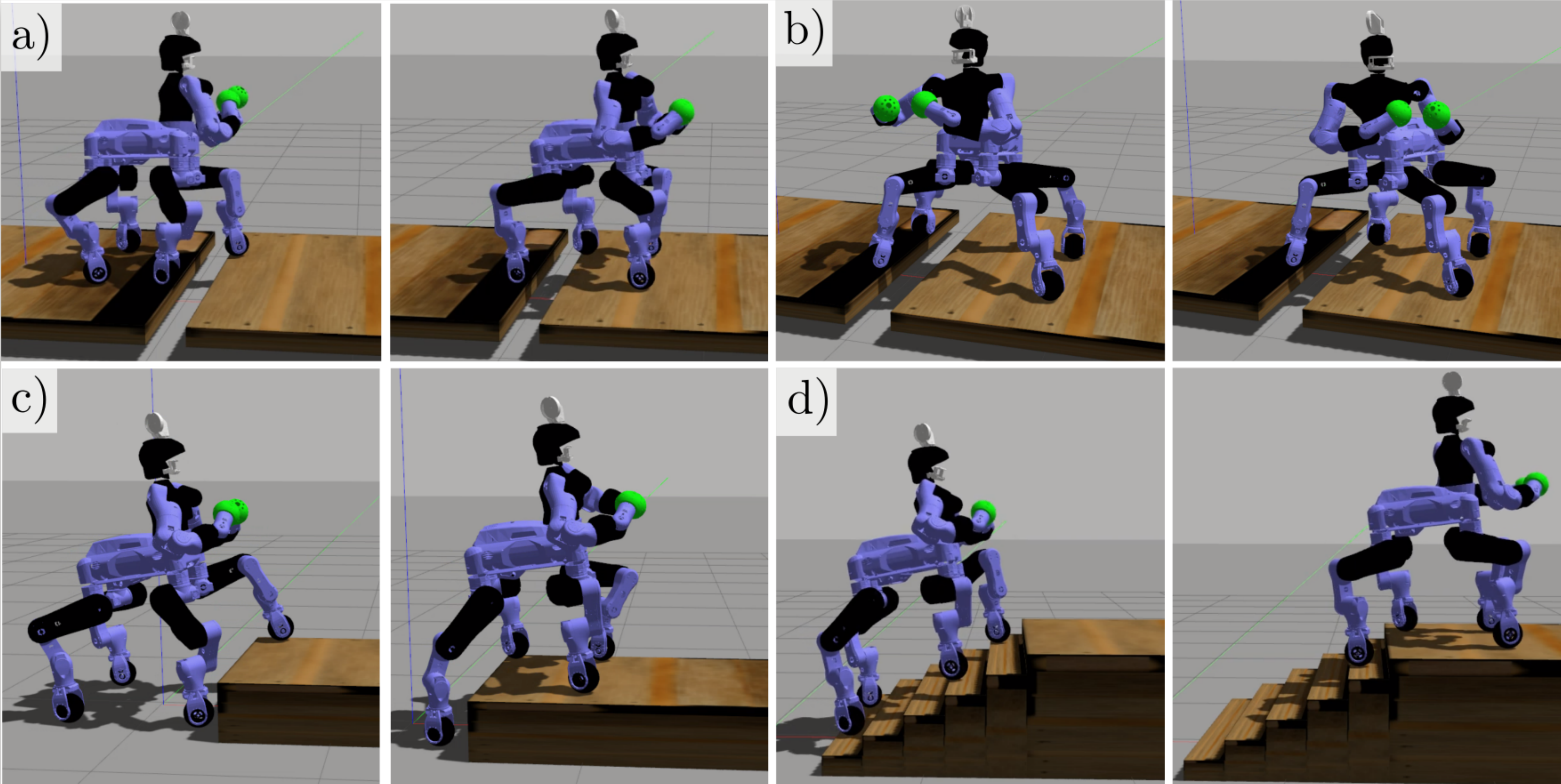
  \caption{CENTAURO negotiating non-flat terrain while carrying 20 kg payload in simulation. a) 0.2 m gap, b) 0.25 m gap, c) 0.3 m height platform and d) staircase. Motion plans from consecutive TOs with 4 step horizon are replayed. The attached green ball EEs consist the attached payloads.}
  \label{Fig:climb}
\end{figure}

In this section the ability of the payload-aware planner to generate plans for efficiently traversing a variety of non-flat terrains is presented. TO with 4 step horizon is run and replayed consecutively on the robot in order to synthesize large motions. In Fig. \ref{Fig:climb}, CENTAURO is shown to negotiate gaps, a 0.3 m height platform (36.8 \% of leg's length) and a staircase comprised of 0.1 m height stairs in simulation. The same platform step up motion is not completed successfully using locomotion-only planning due to fast and large base pitch and yaw motions before the last step (not displayed due to lack of space).
The second part of the accompanying video shows the above complete motions as well as negotiation of inclined terrain. This is the first work to demonstrate a quadruped manipulator climbing such a high step, gaps and a staircase under such heavy payload in simulation. 

\subsection{Implementation and computational efficiency}
\label{sec:computational}

The motion planner is implemented in Python within the symbolic framework of \emph{CasADi} \cite{Andersson2019} and solved with the Ipopt solver \cite{wachter2006implementation} (with custom options ma57 linear solver and adaptive barrier parameter update strategy). NLP discretization is done at 5 Hz. 
The computations presented were run on an Intel Core i9-10900K CPU at 3.70 GHz and all simulations in Gazebo simulator with ODE physics.

As shown in Table \ref{tab:convergence}, the time needed for the optimal solution of the payload-aware planner (P-OFF) is more than 30 times shorter than the planning horizon, namely $\sim400$ ms for 13 sec. (Sc. 1-3) and $\sim1.4$ s for 46 sec. of motion (multiple TOs for the step up of Fig. \ref{Fig:climb}c)), respectively. Compared with the locomotion-only case (L) convergence time is increased since the formulation is augmented with the manipulation mode and both motion and force of arm EEs are optimized, which renders a NLP with higher nonlinearity. Zero initial guess was provided to the solver.

Based on the achieved performance the payload-aware formulation can run in a receding horizon fashion. Continuous walking with the strides of Sc. 1-3 and the step up motion is planned online at 5 Hz with 4 sec. horizon in simulation (included in the last part of the video). The implementation provides insight about the potentiality of the approach for online planning. In this case the solver is warm-started with an initial guess that consists of the previous solution for the common knots and the last available knot solution for the remaining ones. The mean convergence time for a solution is shown in Table \ref{tab:convergence} (P-RH). Although the horizon is of considerable length convergence time is more than 50 times shorter (75 times for the step up motion), which renders future implementations of higher frequency feasible.

\begin{table}
\caption{Locomotion-only (L), Payload-Aware Offline (P-OFF) and Payload-Aware Receding Horizon (P-RH) Mean Convergence Time (iterations) From 5 Samples}
\label{tab:convergence}
\begin{center}
\begin{tabular}{|c||c|c|c|c|}
\hline
 & \emph{Sc. 1} [ms] & \emph{Sc. 2} [ms] & \emph{Sc. 3} [ms] & Step up [s]\\
\hline
\emph{L} &   80.4 (25) & 77.23 (26) &    70.61 (24) & 0.319 (134)\\
\hline
\emph{P-OFF} &   375.34 (44) & 399.44 (46) &    392.93 (45) & 1.41 (194)\\
\hline
\emph{P-RH} &   60.43 (15) & 64.53 (16) &    76.75 (18) & 0.053 (13)\\
\hline
\end{tabular}
\end{center}
\end{table}

%% file: figures/arm_contribution.pdf_tex
\begingroup%
  \makeatletter%
  \providecommand\color[2][]{%
    \errmessage{(Inkscape) Color is used for the text in Inkscape, but the package 'color.sty' is not loaded}%
    \renewcommand\color[2][]{}%
  }%
  \providecommand\transparent[1]{%
    \errmessage{(Inkscape) Transparency is used (non-zero) for the text in Inkscape, but the package 'transparent.sty' is not loaded}%
    \renewcommand\transparent[1]{}%
  }%
  \providecommand\rotatebox[2]{#2}%
  \newcommand*\fsize{\dimexpr\f@size pt\relax}%
  \newcommand*\lineheight[1]{\fontsize{\fsize}{#1\fsize}\selectfont}%
  \ifx\svgwidth\undefined%
    \setlength{\unitlength}{1279.5bp}%
    \ifx\svgscale\undefined%
      \relax%
    \else%
      \setlength{\unitlength}{\unitlength * \real{\svgscale}}%
    \fi%
  \else%
    \setlength{\unitlength}{\svgwidth}%
  \fi%
  \global\let\svgwidth\undefined%
  \global\let\svgscale\undefined%
  \makeatother%
  \begin{picture}(1,0.4038687)%
    \lineheight{1}%
    \setlength\tabcolsep{0pt}%
    \put(0,0){\includegraphics[width=\unitlength,page=1]{arm_contribution.pdf}}%
  \end{picture}%
\endgroup%

%% file: figures/pelvis_trj.pdf_tex
\begingroup%
  \makeatletter%
  \providecommand\color[2][]{%
    \errmessage{(Inkscape) Color is used for the text in Inkscape, but the package 'color.sty' is not loaded}%
    \renewcommand\color[2][]{}%
  }%
  \providecommand\transparent[1]{%
    \errmessage{(Inkscape) Transparency is used (non-zero) for the text in Inkscape, but the package 'transparent.sty' is not loaded}%
    \renewcommand\transparent[1]{}%
  }%
  \providecommand\rotatebox[2]{#2}%
  \newcommand*\fsize{\dimexpr\f@size pt\relax}%
  \newcommand*\lineheight[1]{\fontsize{\fsize}{#1\fsize}\selectfont}%
  \ifx\svgwidth\undefined%
    \setlength{\unitlength}{877.5bp}%
    \ifx\svgscale\undefined%
      \relax%
    \else%
      \setlength{\unitlength}{\unitlength * \real{\svgscale}}%
    \fi%
  \else%
    \setlength{\unitlength}{\svgwidth}%
  \fi%
  \global\let\svgwidth\undefined%
  \global\let\svgscale\undefined%
  \makeatother%
  \begin{picture}(1,0.53675216)%
    \lineheight{1}%
    \setlength\tabcolsep{0pt}%
    \put(0,0){\includegraphics[width=\unitlength,page=1]{pelvis_trj.pdf}}%
  \end{picture}%
\endgroup%

%% file: figures/step-up_stairs_comp.pdf_tex
\begingroup%
  \makeatletter%
  \providecommand\color[2][]{%
    \errmessage{(Inkscape) Color is used for the text in Inkscape, but the package 'color.sty' is not loaded}%
    \renewcommand\color[2][]{}%
  }%
  \providecommand\transparent[1]{%
    \errmessage{(Inkscape) Transparency is used (non-zero) for the text in Inkscape, but the package 'transparent.sty' is not loaded}%
    \renewcommand\transparent[1]{}%
  }%
  \providecommand\rotatebox[2]{#2}%
  \newcommand*\fsize{\dimexpr\f@size pt\relax}%
  \newcommand*\lineheight[1]{\fontsize{\fsize}{#1\fsize}\selectfont}%
  \ifx\svgwidth\undefined%
    \setlength{\unitlength}{1316.25bp}%
    \ifx\svgscale\undefined%
      \relax%
    \else%
      \setlength{\unitlength}{\unitlength * \real{\svgscale}}%
    \fi%
  \else%
    \setlength{\unitlength}{\svgwidth}%
  \fi%
  \global\let\svgwidth\undefined%
  \global\let\svgscale\undefined%
  \makeatother%
  \begin{picture}(1,0.50256411)%
    \lineheight{1}%
    \setlength\tabcolsep{0pt}%
    \put(0,0){\includegraphics[width=\unitlength,page=1]{step-up_stairs_comp.pdf}}%
  \end{picture}%
\endgroup%

%% file: experiments.tex
\section{EXPERIMENTAL VALIDATION AND DISCUSSION OF THE APPROACH}

\subsection{Experimental validation}

\begin{figure}
  \centering
  \def\svgwidth{\columnwidth}
  \graphicspath{{figures/}}
  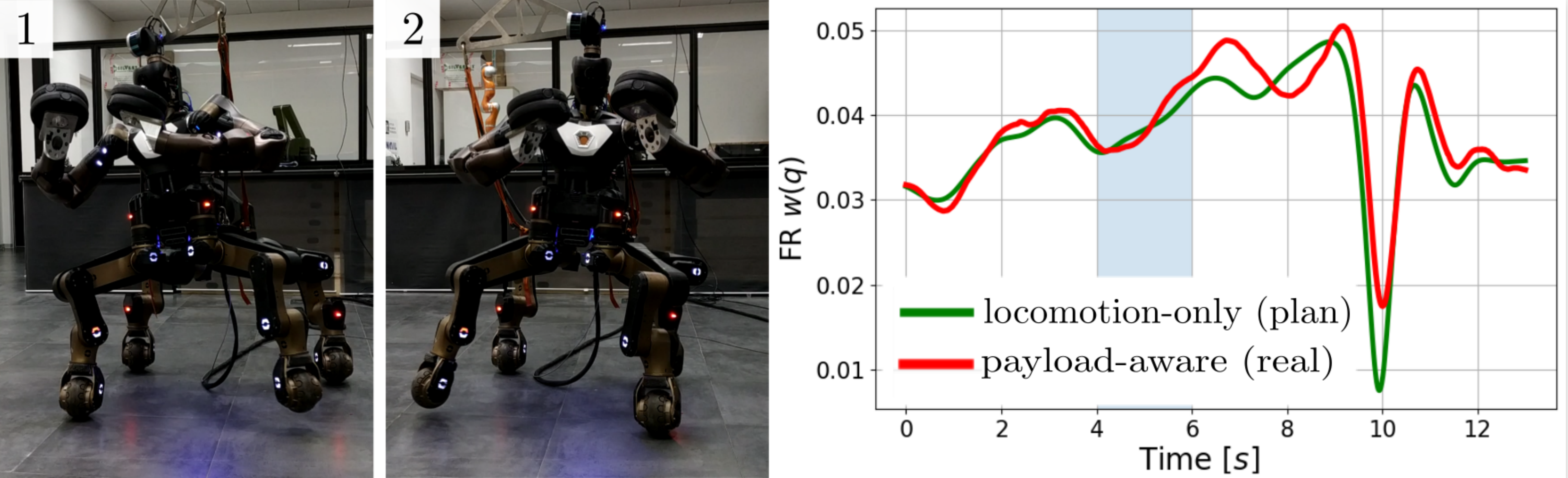
  \caption{Snapshots of lateral walking experiments on flat terrain (left) FR leg manipulability comparison based on experimental and simulation data on the payload-aware and locomotion-only cases, respectively. (right)}
  \label{Fig:experiment}
\end{figure}

\begin{figure}
  \centering
  \def\svgwidth{\columnwidth}
  \graphicspath{{figures/}}
  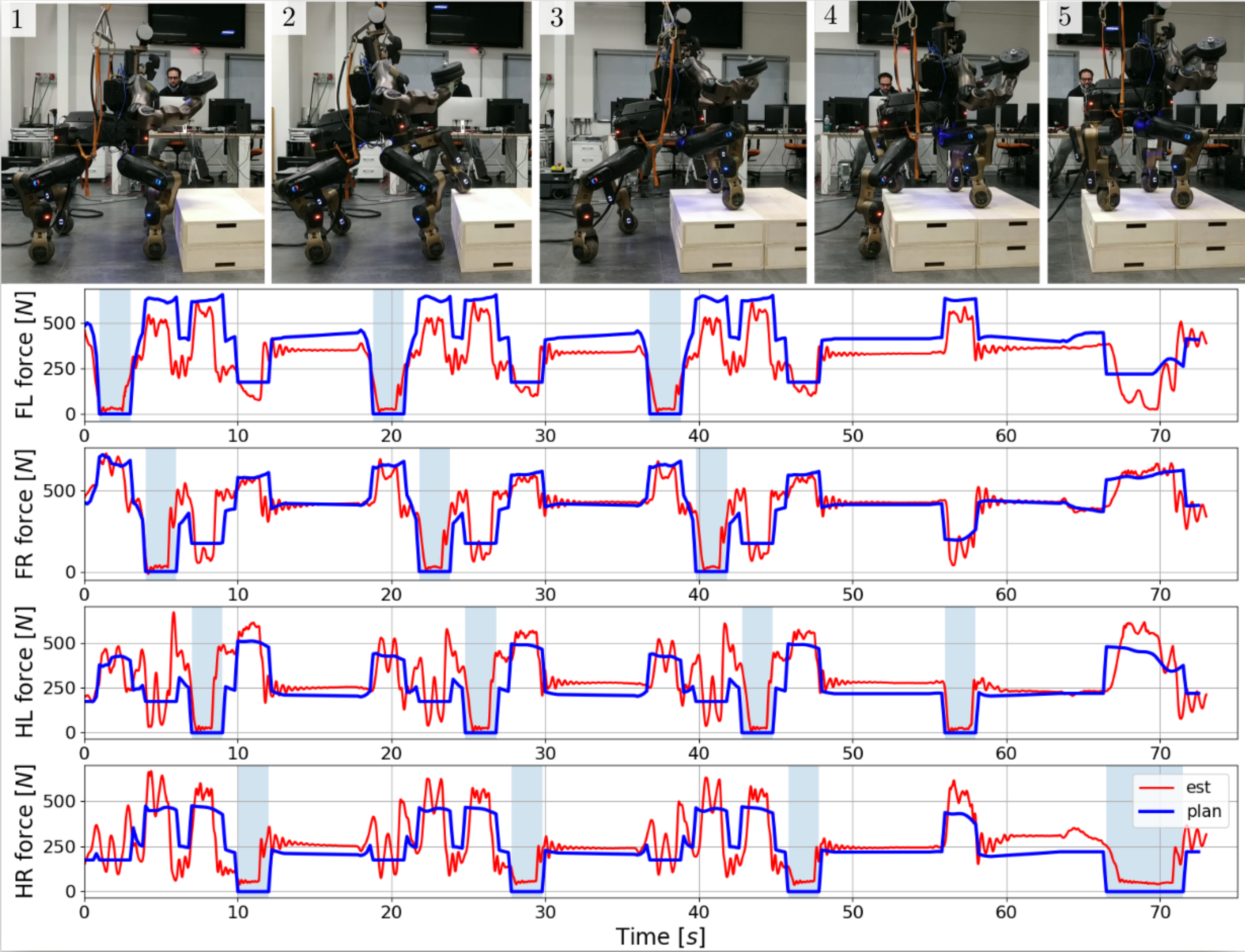
  \caption{Snapshots of CENTAURO stepping up on a 0.3 m height platform (top), planned (blue) and estimated (red) normal force component at feet EEs (bottom). Shaded regions denote swing periods.}
  \label{Fig:experiment_stepup}
\end{figure}

The efficiency on the real hardware is showcased with experiments on the CENTAURO robot carrying 8.5 kg payload attached at each arm (85 \% of each arm's payload capacity, total 15.1 \% of the robot mass). The scenarios include motions of 4 lateral steps on flat terrain as well as stepping up on a 0.3 m height platform (through multiple offline TOs), which are shown in Fig. \ref{Fig:experiment} and \ref{Fig:experiment_stepup}, respectively. 
The motions can be found in the third part of the accompanying video. It is noted that the used boxes in \eqref{eq:box_ee} are set more conservative than in simulation for safety reasons. Moreover, a larger stability threshold $f^z_{min}=175$ N in \eqref{eq:unilateral} is used to increase balance robustness against the sim-to-real gap.

The data from the lateral stepping experiment of Fig. \ref{Fig:experiment} indicate that the generated trajectories are accurately tracked from the real hardware such that the real FR leg manipulability remains higher than the one planned in the locomotion-only case. 
In Fig. \ref{Fig:experiment_stepup} the planned and estimated normal force components during the platform stepping up experiment are depicted for the feet EEs. The estimated forces follow in general the trend of the planned ones. The presented force tracking errors are mainly due to the fact that forces are not explicitly tracked, joint position control is used (there is force redundancy) and there are estimation errors. Finally, the estimated force components in Fig. \ref{Fig:experiment_stepup} often reach low values at each leg when the one diagonal to it is swinging due to the momentum produced by the joint velocities when swinging fast a robot leg. Nevertheless the robot remains stable due to the considered stability constraint \eqref{eq:unilateral}.

\subsection{Discussion}

This work uses simplified robot and payload models to achieve computational efficiency for the high dimensional CENTAURO robot. Nevertheless, the performance obtained in simulation and real hardware indicate that the model assumptions are not restrictive for the variety of showcased motions. 
More challenging motions (larger singularity-free strides, complex and fast maneuvers as well as more challenging terrain) will require more accurate enforcement of kinematic constraints (e.g. accurate self-collision avoidance, arm EE workspaces) and reasoning about richer models. Complete kinematics would also enable better exploitation of the arm EEs workspace and, thus, increase the manipulation contribution. To the best of our knowledge, there is no work efficiently reasoning about full models of a robot with dimension equal or greater than CENTAURO.

%% file: figures/experiments_comp.pdf_tex
\begingroup%
  \makeatletter%
  \providecommand\color[2][]{%
    \errmessage{(Inkscape) Color is used for the text in Inkscape, but the package 'color.sty' is not loaded}%
    \renewcommand\color[2][]{}%
  }%
  \providecommand\transparent[1]{%
    \errmessage{(Inkscape) Transparency is used (non-zero) for the text in Inkscape, but the package 'transparent.sty' is not loaded}%
    \renewcommand\transparent[1]{}%
  }%
  \providecommand\rotatebox[2]{#2}%
  \newcommand*\fsize{\dimexpr\f@size pt\relax}%
  \newcommand*\lineheight[1]{\fontsize{\fsize}{#1\fsize}\selectfont}%
  \ifx\svgwidth\undefined%
    \setlength{\unitlength}{1316.25bp}%
    \ifx\svgscale\undefined%
      \relax%
    \else%
      \setlength{\unitlength}{\unitlength * \real{\svgscale}}%
    \fi%
  \else%
    \setlength{\unitlength}{\svgwidth}%
  \fi%
  \global\let\svgwidth\undefined%
  \global\let\svgscale\undefined%
  \makeatother%
  \begin{picture}(1,0.30484329)%
    \lineheight{1}%
    \setlength\tabcolsep{0pt}%
    \put(0,0){\includegraphics[width=\unitlength,page=1]{experiments_comp.pdf}}%
  \end{picture}%
\endgroup%

%% file: figures/experiment_stepup_comp.pdf_tex
\begingroup%
  \makeatletter%
  \providecommand\color[2][]{%
    \errmessage{(Inkscape) Color is used for the text in Inkscape, but the package 'color.sty' is not loaded}%
    \renewcommand\color[2][]{}%
  }%
  \providecommand\transparent[1]{%
    \errmessage{(Inkscape) Transparency is used (non-zero) for the text in Inkscape, but the package 'transparent.sty' is not loaded}%
    \renewcommand\transparent[1]{}%
  }%
  \providecommand\rotatebox[2]{#2}%
  \newcommand*\fsize{\dimexpr\f@size pt\relax}%
  \newcommand*\lineheight[1]{\fontsize{\fsize}{#1\fsize}\selectfont}%
  \ifx\svgwidth\undefined%
    \setlength{\unitlength}{922.5bp}%
    \ifx\svgscale\undefined%
      \relax%
    \else%
      \setlength{\unitlength}{\unitlength * \real{\svgscale}}%
    \fi%
  \else%
    \setlength{\unitlength}{\svgwidth}%
  \fi%
  \global\let\svgwidth\undefined%
  \global\let\svgscale\undefined%
  \makeatother%
  \begin{picture}(1,0.76504065)%
    \lineheight{1}%
    \setlength\tabcolsep{0pt}%
    \put(0,0){\includegraphics[width=\unitlength,page=1]{experiment_stepup_comp.pdf}}%
  \end{picture}%
\endgroup%

%% file: Conclusion.tex
\section{CONCLUSION AND FUTURE WORK}

This paper proposes an efficient TO formulation that deploys locomotion and payload manipulation of quadruped mobile manipulators in heavy payload transportation tasks. The framework demonstrates enhanced performance on flat and non-flat terrain under payload that exceeds 15 \% of the robot's mass and overcomes the locomotion-only approach that is proved to be poor for kinematically demanding motions. This work highlights the advantage gain on planning tasks that involve large physical interaction by considering both locomotion and manipulation of quadruped manipulators together from the planning stage. 

Future work shall focus on online planning for the real robot with perception in the loop. Considering more accurate models for CENTAURO robot is an interesting, yet challenging direction. Finally, picking up/placing down heavy payloads consists future work as well.

%% file: references.tex
\bibliographystyle{IEEEtran}
\bibliography{bibliography}